\documentclass[11pt]{article}

\usepackage{acl}

\usepackage[T1]{fontenc}
\usepackage[utf8]{inputenc}
\usepackage{times}
\usepackage{latexsym}
\usepackage{microtype}
\usepackage{inconsolata}

\usepackage{siunitx}
\usepackage{multirow}

\newcolumntype{I}{S[table-format=2.0]}
\newcolumntype{F}{S[table-format=1.3]}
\newcolumntype{P}{S[table-format=2.3]}

\usepackage{graphicx}
\usepackage{subcaption}
\usepackage{booktabs}
\usepackage{multirow}
\usepackage{amsmath, amssymb, mathtools, amsthm, bm}
\usepackage{nicefrac}
\usepackage{physics2}
\usephysicsmodule{ab}
\usepackage{adjustbox}
\usepackage{array}
\usepackage[capitalize,noabbrev]{cleveref}

\allowdisplaybreaks[2] 
\newenvironment{fitpage}{\begin{adjustbox}{max width=\textwidth}}{\end{adjustbox}}

\DeclareMathOperator*{\argmin}{arg\,min}

\newcommand{\RR}{\mathbb{R}}
\newcommand{\EE}{\mathbb{E}}
\newcommand{\DD}{\mathcal{D}}

\newcommand{\dFR}{d_{\mathrm{FR}}}
\newcommand{\KL}{\mathrm{KL}}

\newcommand{\Logm}[1]{\operatorname{Log}_{#1}}
\newcommand{\Expm}[1]{\operatorname{Exp}_{#1}}
\newcommand{\defeq}{\coloneq}

\newcommand{\best}[1]{\textbf{#1}}

\newcommand{\modelname}{\textsc{KARCHER}}

\title{Functionality-Oriented LLM Merging on the Fisher--Rao Manifold}
\author{
Jiayu Wang \\
Pennsylvania State University \\
\texttt{garion@psu.edu} \\
\And
Zuojun Ye\thanks{Independent developer. GitHub: \url{https://github.com/win10ogod}} \\
Independent Developer \\
\texttt{jmes100010@gmail.com} \\
\And
Wenpeng Yin \\
Pennsylvania State University \\
\texttt{wenpeng@psu.edu} \\
}

\begin{document}
\maketitle

\begin{abstract}
Weight-space merging aims to combine multiple fine-tuned LLMs into a single model without retraining, yet most existing approaches remain fundamentally \emph{parameter-space} heuristics. This creates three practical limitations. First, linear averaging, task vectors, and related rules operate on Euclidean coordinates, even though \textit{the desired goal is to merge \emph{functionality}---i.e., predictive behaviors---across tasks}. Second, when the source checkpoints are farther apart or more heterogeneous, Euclidean blends often trigger \emph{representation collapse}, manifested as activation variance shrinkage and effective-rank degradation, which sharply degrades accuracy. Third, many geometry-inspired methods are most natural for \emph{two-model} interpolation (e.g., SLERP-style rules) and do not extend cleanly to merging $N>2$ experts with a principled objective. We address these issues by formulating model merging as computing a (weighted) Karcher/Fr\'echet mean on the Fisher--Rao manifold, which is locally equivalent to minimizing a KL-based \emph{function distance} between predictive distributions. We derive a practical fixed-point algorithm using a lightweight spherical proxy that preserves norms and generalizes directly to multi-expert merging. Across various benchmarks  and collapse diagnostics, our method remains stable as the number and heterogeneity of merged models increase, consistently outperforming prior baselines.\footnote{
Code: \url{https://github.com/arcee-ai/mergekit/commit/09bbb0ae282c6356567f05fe15a28055b9dc9390}.
Our implementation builds on MergeKit \citep{goddard2025arceesmergekittoolkitmerging}, the MergeKit repository is not authored by us, but the pull request that modifies
\texttt{karcher.py} is our contribution.}
\end{abstract}

\section{Introduction}
Model merging aims to combine capabilities from multiple fine-tuned LLMs into a single
model \emph{without} additional training. In practice, naive Euclidean operations
(e.g., averaging weights or task vectors) can lead to \emph{function mismatch} and
\emph{collapse}: merged representations become weakly input-dependent (variance
collapse) and the effective dimensionality of activations degrades (rank collapse),
hurting accuracy and perplexity \citep{jordan2023repair,qu2025vanishing,skorobogatov2025subspace,sharma2024nonlocal}.
A geometric explanation is that low-loss regions form curved valleys; fine-tuned
checkpoints often lie on thin shells around a base model, and linear blends shrink
norms and drift off the high-performing manifold \citep{jang2024modelstock}.

\paragraph{From parameter chords to function distance.}
A principled notion of distance between models is the discrepancy between their
\emph{predictive distributions}. For small parameter displacements,
the Fisher--Rao (FR) metric links parameter-space geometry to distribution-space
divergence:
\begin{equation}
\begin{aligned}
  \dFR^2(\bm\theta,\bm\theta')
  &\approx (\bm\theta-\bm\theta')^\top
  \mathbf{F}(\bm\theta)\,(\bm\theta-\bm\theta') \\
  &\approx 2\,\KL\!\left(p_{\bm\theta}\,\|\,p_{\bm\theta'}\right),
\end{aligned}
\label{eq:fr_kl_local}
\end{equation}
where $\mathbf{F}(\bm\theta)$ is the Fisher information matrix and the approximation
holds locally. This motivates merging by minimizing an FR-based barycentric objective,
which corresponds to minimizing a KL-based \emph{function distance}. Concretely, for a
task distribution $\DD^{(i)}$ and a teacher model $\bm\theta^{(i)}$,
\begin{equation}
\begin{aligned}
  &\EE_{(x,y)\sim\DD^{(i)}}\![-\log p_{\bm\theta}(y\mid x)]
   \\
  &= \text{const} + \EE_{x\sim\DD^{(i)}}\!\Big[
      \KL\!\Big(
        p_{\bm\theta^{(i)}}(\cdot\mid x)\,\|\,p_{\bm\theta}(\cdot\mid x)
      \Big)
    \Big].
\end{aligned}
\label{eq:nll_decomp}
\end{equation}
so reducing the expected KL-to-teachers aligns with improving NLL/PPL.

\paragraph{Why Karcher means help more when models are farther apart.}
A key geometric point (often glossed over in practice) is that the difference between
a straight chord and the true geodesic \emph{grows with distance and curvature}. When
the source models are close (small task vectors / mild fine-tuning), many merge rules
behave similarly because the manifold is nearly flat locally.
However, when models are farther apart---e.g., larger fine-tuning deltas, more
heterogeneous experts, or simply merging more models---Euclidean averaging cuts across
curvature, exacerbating norm shrinkage and interference. In this regime, a geodesic
barycenter (Karcher mean) is typically more advantageous, because it remains on (or
near) the high-performing manifold that connects the experts.

Overall, the contributions of this work is threefold: i) We formulate model merging as computing a (weighted) Karcher/Fr\'echet mean on
        the Fisher--Rao manifold, directly targeting KL-based function distance; ii) We derive a practical fixed-point algorithm with a lightweight spherical proxy
        that (i) reduces to SLERP for two-model merges and (ii) scales to $N>2$ models; iii) We provide empirical evidence that the proposed merge is stable under
        increasing merge scale and heterogeneity, and mitigates collapse diagnostics
        compared to strong baselines.

\section{Related Work}
\subsection{Weight-space merging}
\paragraph{Linear/task-vector merges.}
Model soups and task arithmetic average weights or deltas relative to a base, but can
be sensitive to misalignment and interference
\citep{wortsman2022soups,ainsworth2022gitrebasing}. TIES
\citep{yadav2023ties} trims small updates and resolves sign conflicts; DARE
\citep{yu2023dare} drops and rescales sparse deltas; DELLA
\citep{deep2024della} uses magnitude-aware sampling. These methods are effective in
many settings but remain Euclidean heuristics that can become brittle as models become
more diverse.

\subsection{Geometric and Fisher-inspired views}
\paragraph{Two-model geodesics.}
SLERP preserves norm on a hypersphere and often outperforms linear interpolation for
two models \citep{wortsman2022wiseft}. ChipAlign applies geodesic interpolation for
instruction alignment in domain LLMs \citep{deng2024chipalign}. Model Stock highlights
thin-shell geometry and proposes center-of-shell averaging across seeds/checkpoints
\citep{jang2024modelstock}. These ideas motivate geodesic reasoning, but are either
specialized to two models (SLERP/ChipAlign) or rely on specific shell structures.

\paragraph{Fisher weighting.}
Fisher-weighted averaging merges models by weighting parameters according to Fisher
information \citep{DBLPMatenaR22}. Our approach is complementary: rather than
performing a Fisher-weighted \emph{Euclidean} average, we compute a (proxy) Riemannian
barycenter motivated by Fisher--Rao geometry.

\begin{table*}[t]
  \centering
  \small
  \setlength{\tabcolsep}{4.2pt}
  \renewcommand{\arraystretch}{0.98}
\begin{tabular}{l|ccccc|c}
  \toprule
  Method & HellaSwag & BBH & MMLU-Pro & MuSR & GPQA-D & Avg \\
  \midrule
  \multicolumn{7}{c}{\textbf{Merging $m=2$ models}} \\
  \midrule
  (Multi-)SLERP   & 0.825 & 0.640 & 0.523 & 0.500 & 0.386 & 0.575 \\
  DARE-LERP       & 0.770 & 0.530 & 0.418 & 0.460 & 0.325 & 0.501 \\
  DARE-TIES       & 0.767 & 0.542 & 0.434 & 0.513 & 0.330 & 0.517 \\
  DELLA-LERP      & 0.766 & 0.547 & 0.432 & 0.492 & 0.320 & 0.511 \\
  DELLA-TIES      & 0.765 & 0.541 & 0.426 & 0.487 & 0.325 & 0.509 \\
  LERP            & 0.825 & 0.643 & 0.524 & 0.504 & 0.391 & 0.577 \\
  TIES            & 0.799 & 0.590 & 0.471 & 0.505 & 0.335 & 0.540 \\
  Arcee Fusion    & 0.777 & 0.611 & 0.490 & 0.475 & 0.335 & 0.538 \\
  \modelname~(ours) & \best{0.830} & \best{0.653} & \best{0.532} & \best{0.523} & \best{0.448} & \best{0.597} \\
  \midrule
  \multicolumn{7}{c}{\textbf{Merging $m=5$ models}} \\
  \midrule
  (Multi-)SLERP   & 0.244 & 0.280 & 0.105 & 0.356 & 0.209 & 0.239 \\
  DARE-LERP       & 0.259 & 0.276 & 0.108 & 0.364 & 0.189 & 0.239 \\
  DARE-TIES       & 0.252 & 0.283 & 0.113 & 0.332 & 0.209 & 0.238 \\
  DELLA-LERP      & 0.253 & 0.281 & 0.113 & 0.345 & 0.214 & 0.241 \\
  DELLA-TIES      & 0.266 & 0.269 & 0.111 & 0.348 & 0.244 & 0.248 \\
  LERP            & 0.811 & 0.613 & 0.468 & 0.499 & 0.320 & 0.542 \\
  Model Stock     & 0.823 & 0.630 & 0.508 & 0.468 & 0.355 & 0.557 \\
  SCE             & 0.735 & 0.529 & 0.342 & 0.442 & 0.260 & 0.461 \\
  TIES            & 0.271 & 0.288 & 0.109 & 0.357 & 0.239 & 0.253 \\
  \modelname~(ours) & \best{0.836} & \best{0.680} & \best{0.558} & \best{0.532} & \best{0.443} & \best{0.610} \\
  \bottomrule
\end{tabular}
  \caption{Results when merging $m=2$ or $m=5$ LLMs. All metrics are normalized to the $[0,1]$ scale. HellaSwag/BBH/MuSR use \texttt{acc\_norm}. MMLU-Pro and GPQA-D are reported as normalized accuracies. Avg is the mean over all five tasks.}
  \label{tab:mainresults}
  \vspace{-3mm}
\end{table*}

\section{Method: Fisher--Rao Karcher mean merging}

\subsection{Notation}
Let $\bm{\theta}^{(0)}\in\RR^d$ denote base (pretrained) parameters; experts
$\{\bm{\theta}^{(i)}\}_{i=1}^N$ are fine-tuned variants; task vectors are
$\bm{\delta}^{(i)}\defeq\bm{\theta}^{(i)}-\bm{\theta}^{(0)}$; mixture weights
$\alpha^{(i)}\geq 0$ with $\sum_i\alpha^{(i)}=1$. For an input $x$ and label $y$, the
predictive distribution is $p_{\bm{\theta}}(y\mid x)$. The Fisher information
$\mathbf{F}_{\bm{\theta}}$ induces the Fisher--Rao geodesic distance $\dFR(\cdot,\cdot)$;
$\Logm{\bm\theta}(\cdot)$ and $\Expm{\bm\theta}(\cdot)$ denote Riemannian log/exp maps.

\subsection{Objective}
Given experts $\{\bm\theta^{(i)}\}_{i=1}^N$ and weights $\alpha^{(i)}$, we target the
Fr\'echet/Karcher mean on the Fisher--Rao manifold:
\begin{equation}
  \bm\theta^*
  \;\defeq\;
  \argmin_{\bm\theta}\;
  \sum_{i=1}^N \alpha^{(i)}\, \dFR^2\!\left(\bm\theta,\bm\theta^{(i)}\right).
  \label{eq:karcher_obj}
\end{equation}
At an optimum (under mild conditions), the weighted Riemannian first-order condition
is
\begin{equation}
  \sum_{i=1}^N \alpha^{(i)}\, \Logm{\bm\theta^*}\!\left(\bm\theta^{(i)}\right)
  \;=\; \bm 0.
  \label{eq:karcher_stationary}
\end{equation}
Intuitively, \cref{eq:karcher_obj} minimizes the average geodesic distance between the
merged model and all experts; via \cref{eq:fr_kl_local}, this corresponds to minimizing
a KL-based function distance.

\subsection{Fixed-point iteration}
A standard approach to computing Karcher means is a fixed-point update (equivalently,
a Riemannian gradient step for \cref{eq:karcher_obj}):
\begin{equation}
\begin{aligned}
  \bm v^{(t)}
  &= \sum_{i=1}^N \alpha^{(i)}\,
  \Logm{\bm\theta^{(t)}}\!\left(\bm\theta^{(i)}\right), \\
  \bm\theta^{(t+1)}
  &= \Expm{\bm\theta^{(t)}}\!\left(\eta\,\bm v^{(t)}\right),
\end{aligned}
\label{eq:karcher_update}
\end{equation}
with step size $\eta\in(0,1]$.
For a two-model merge between $\bm\theta^{(0)}$ and $\bm\theta^{(1)}$ with equal
weights, initializing at $\bm\theta^{(0)}$ yields
$\bm\theta^{(1)} = \Expm{\bm\theta^{(0)}}(\tfrac{1}{2}\Logm{\bm\theta^{(0)}}(\bm\theta^{(1)}))$,
i.e., a geodesic midpoint. Under a spherical proxy (below), this reduces to SLERP.

\subsection{Practical approximation: spherical proxy with norm preservation}
Computing exact Fisher--Rao log/exp maps for modern LLMs is intractable. We adopt a
proxy motivated by two empirical observations from prior analyses: (i) fine-tuned
checkpoints often lie on a thin shell around the base model, and (ii) norm shrinkage
is a major failure mode of Euclidean interpolation \citep{jang2024modelstock}.

\paragraph{Spherical Karcher mean (directional barycenter).}
We treat each parameter block (e.g., layer or tensor group) as a vector and normalize
it to the unit sphere. We then compute the Karcher mean on $S^{d-1}$ using the
closed-form log/exp maps on the sphere, and finally rescale by a representative norm
(e.g., the mean norm of sources for that block). This yields a \emph{norm-preserving}
merge that captures a first-order notion of curved geometry while remaining extremely
lightweight.

\paragraph{Connection to Fisher geometry.}
Locally, Fisher information weights directions that strongly affect the predictive
distribution \citep{DBLPMatenaR22}. In practice, we implement the update
blockwise, and can incorporate diagonal/KFAC Fisher estimates as a
natural-gradient-style preconditioning inside the log map approximation. This protects
high-Fisher directions and reduces destructive interference in sensitive subspaces.

\paragraph{Why this mitigates collapse.}
Variance/rank collapse is associated with merges drifting toward bias-dominated or
low-dimensional regimes \citep{jordan2023repair,qu2025vanishing,skorobogatov2025subspace}.
By minimizing a KL-weighted barycentric objective, the Karcher update keeps the merged
predictive distribution close to \emph{all} experts. Geometrically, the update follows
a geodesic-like path that avoids chordal shortcuts responsible for norm shrinkage and
feature disappearance.

\section{Experiments}
\subsection{Settings}
We evaluate on the following benchmarks:
\textbf{GPQA-Diamond} \citep{rein2023gpqa} (\texttt{acc\_norm}),
\textbf{HellaSwag} \citep{zellers2019hellaswag} (\texttt{acc\_norm}),
\textbf{MMLU-Pro} \citep{wang2024mmlupro} (\texttt{5-shot acc}),
\textbf{MuSR} \citep{sprague2023musr} (\texttt{acc\_norm}),
and \textbf{BBH} \citep{suzgun2022bbh},
plus the unweighted \textbf{Avg}.
All evaluations use the LM Evaluation Harness \citep{eval-harness} with default seed.

\paragraph{Models and merge scale.}
Unless otherwise noted, all merges are performed within the Qwen2.5 family \citep{yang2024qwen25},
so models at a given scale share the same tokenizer and architecture.
We report results in two regimes:
(i) \emph{Pairwise} merges (e.g., base $\leftrightarrow$ instruct) across multiple model sizes; and
(ii) \emph{Multi-expert} merges on Qwen2.5-14B, where we progressively merge
$m\in\{2,\dots,11\}$ models from a pool of Qwen2.5-14B-compatible checkpoints.
\footnote{
HuggingFace model IDs used in the 14B pool:
\texttt{Qwen/Qwen2.5-14B} \citep{qwen2.5},
\texttt{Qwen/Qwen2.5-14B-Instruct-1M} \citep{qwen2.5-1m,qwen2.5},
\texttt{Qwen/Qwen2.5-Coder-14B-Instruct} \citep{qwen2.5},
\texttt{Krystalan/DRT-14B} \citep{wang2024drt},
\texttt{deepseek-ai/DeepSeek-R1-Distill-Qwen-14B} \citep{deepseekai2025deepseekr1incentivizingreasoningcapability},
\texttt{nvidia/OpenReasoning-Nemotron-14B},
\texttt{deepcogito/cogito-v1-preview-qwen-14B},
\texttt{arcee-ai/SuperNova-Medius},
\texttt{netease-youdao/Confucius-o1-14B} \citep{confucius-o1},
\texttt{sthenno-com/miscii-14b-0218} \citep{sthenno_2025},
\texttt{prithivMLmods/Galactic-Qwen-14B-Exp2}.}

\paragraph{Baselines.}
We compare against widely used merge methods, implemented via MergeKit
\citep{DBLP:conf/emnlp/GoddardSEMKBMS24}:
\textbf{Lerp} and \textbf{(Multi-)Slerp} \citep{wortsman2022wiseft},
\textbf{Model Stock} \citep{jang2024modelstock},
\textbf{Ties} \citep{yadav2023ties},
\textbf{DARE-Lerp/Ties} \citep{yu2023dare},
\textbf{DELLA-Lerp/Ties} \citep{deep2024della},
\textbf{SCE} \citep{wan2024fusechat},
and \textbf{Arcee Fusion} \citep{DBLP:conf/emnlp/GoddardSEMKBMS24} (where applicable).
Unless otherwise noted, all merges use equal source weights; for two-way SLERP we use $t=0.5$.

\begin{figure}[t]
  \centering
  \includegraphics[width=\linewidth]{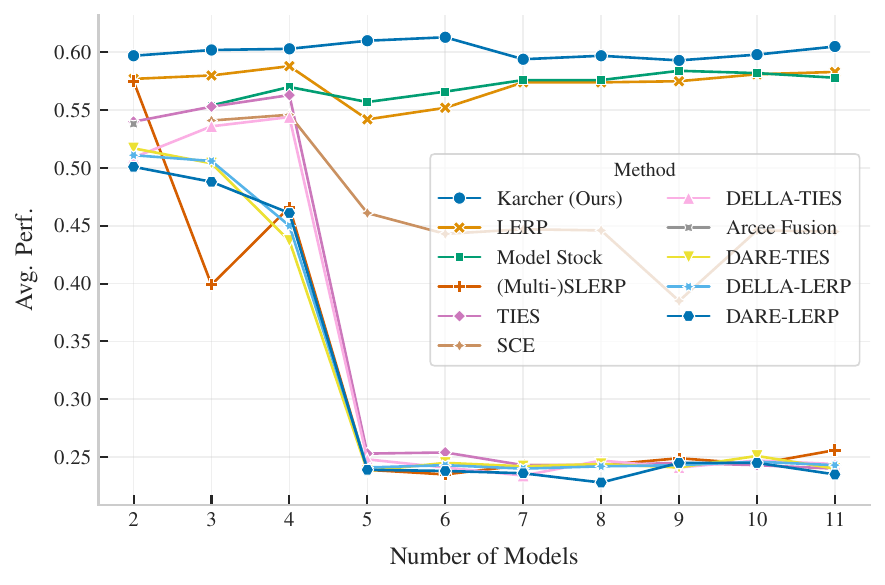}
  \caption{
    Average performance versus the number of merged models $m$.
    As $m$ increases (and the merged set becomes more heterogeneous/farther apart),
    several Euclidean-rule baselines exhibit abrupt collapse around $m\approx 5$,
    remaining in a low-performance regime thereafter.
    The proposed Karcher merge remains stable across $m\in\{2,\dots,11\}$ and achieves
    the best overall performance.
  }
  \label{fig:avg_perf_vs_m}
\end{figure}
\begin{table}[t]
  \centering
      \begin{tabular}{lccc}
        \toprule
        Method                  & 135M         & 360M         & 1.7B         \\
        \midrule
        Ties                    & 0.240        & 0.271        & 0.391        \\
        Slerp                   & 0.230        & 0.274        & 0.395        \\
        Lerp                    & 0.245        & 0.269        & 0.398        \\
       \modelname~(ours) & \best{0.246} & \best{0.282} & \best{0.401} \\
        \bottomrule
      \end{tabular}
  \caption{Comparison across LLM scales (when $m=2$). Scores are normalized in $[0,1]$.}
  \label{tab:compare_scales}
\end{table}

\begin{figure}[t]
\centering

\begin{subfigure}{0.9\linewidth}
    \centering
    \includegraphics[width=\linewidth]{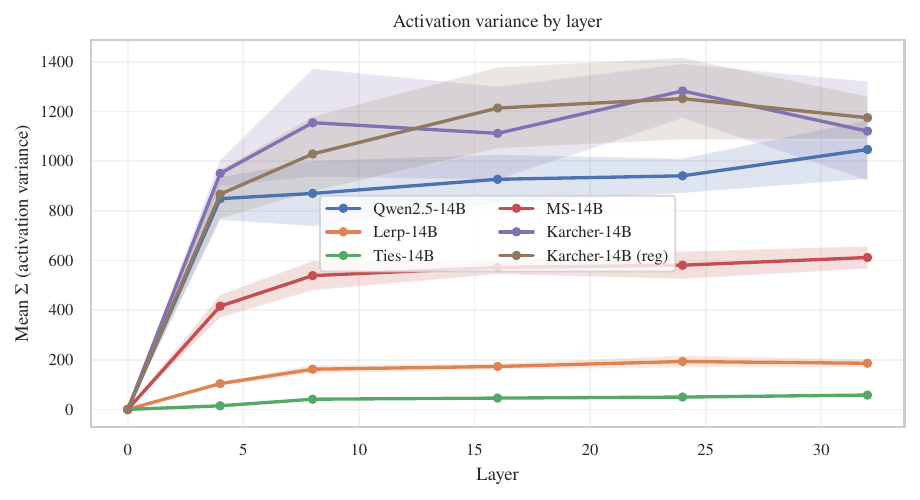}
    \caption{Activation variance across layers.}
    \label{fig:variance_layers}
\end{subfigure}

\vspace{0.5em}

\begin{subfigure}{0.9\linewidth}
    \centering
    \includegraphics[width=\linewidth]{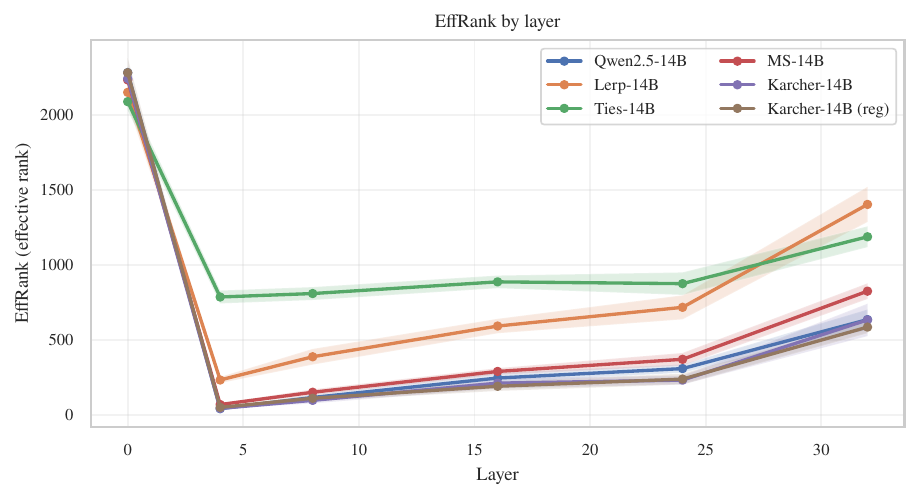}
    \caption{Effective rank across layers.}
    \label{fig:rank_layers}
\end{subfigure}

\caption{
Layerwise diagnostics of activation statistics.
Top: mean activation variance across transformer layers.
Bottom: effective rank (EffRank) of the activation covariance.
Compared with interpolation-based merges (e.g., Lerp and Ties),
Karcher merging preserves both variance and effective dimensionality
across mid-to-deep layers, indicating reduced representation collapse.
}
\label{fig:layerwise_diagnostics}
\end{figure}

\subsection{Results \& Analysis}

We address four evaluation questions.

\paragraph{$\mathcal{Q}_1$: How does \modelname~compare to baseline methods across benchmarks?}
Table~\ref{tab:mainresults} reports detailed performance when merging $m=2$ and $m=5$ LLMs. 
\modelname~consistently outperforms all baselines. Moreover, its advantage becomes more pronounced as $m$ increases (particularly at $m=5$), motivating a closer examination of scalability with respect to the number of merged models (i.e., the next question $\mathcal{Q}_2$).

\paragraph{$\mathcal{Q}_2$: Does \modelname~remain effective when merging more than two LLMs?}
Most baselines are primarily studied and reported in the pairwise ($m=2$) setting,
leaving their multi-model scalability unclear or unstable.
Figure~\ref{fig:avg_perf_vs_m} compares performance from $m=2$ to $m=11$. 
\modelname~remains stable as $m$ grows, whereas several baselines degrade sharply. 
This supports the core geometric claim: geodesic barycenters are most beneficial when
sources are farther apart or more heterogeneous, precisely where chord-based averages
become unreliable.

\paragraph{$\mathcal{Q}_3$: Is \modelname~robust when merging models of different scales?}
Table~\ref{tab:compare_scales} presents pairwise merging across three scales (135M, 360M, and 1.7B). 
Even in this relatively \emph{nearby} regime (two related checkpoints, $m=2$), Karcher remains superior, with a modest gain as expected when geometric discrepancies between models are limited.

\paragraph{$\mathcal{Q}_4$: Can \modelname~relieve the variance and rank collapse problem?}
A common failure mode of interpolation-based merging is that internal activations lose
diversity (variance collapse) and become effectively low-rank (rank collapse)
\citep{jordan2023repair,qu2025vanishing,sharma2024nonlocal}. We report layerwise
activation variance and rank-related diagnostics in Figure \ref{fig:layerwise_diagnostics} (please refer to \cref{tab:layerwise-rank-diagnostics}
in Appendix for more detailed report). Across layers, Karcher-based merges preserve substantially larger effective
rank (EffRank) and numerical rank (NumRank) than interpolation baselines (e.g., Lerp
and Ties), especially in mid-to-deep layers. 

\section{Conclusion}
We formulate model merging as computing a Karcher mean on (a proxy of) the Fisher--Rao
manifold, yielding a geometry-aware merge that minimizes KL-based function distance
rather than Euclidean chord length. The resulting algorithm (i) generalizes SLERP from
two models to $N>2$ models via a principled barycentric objective, (ii) is lightweight
and tuning-light, and (iii) empirically improves stability and average performance
over strong baselines while mitigating collapse diagnostics. Importantly, the benefit
of Karcher merging is most pronounced in the regime where models are farther apart or
more heterogeneous---exactly where Euclidean merging is most prone to failure.

\section*{Limitations}
Our method relies on approximations to Fisher--Rao geometry. In particular, we use a
spherical proxy (plus optional blockwise Fisher preconditioning) rather than exact
Fisher--Rao geodesics, and this proxy may deviate from the true metric in highly
nonlinear regions of the loss landscape. The fixed-point iteration may depend on
initialization, step size, and stopping criteria; we do not provide global convergence
guarantees for arbitrary expert sets. Empirically, our evaluations focus on a
leaderboard-style suite and a limited set of architectures/checkpoints; results may
not fully transfer to other model families, modalities, or highly adversarial
heterogeneous pools. Finally, as with other weight-space merging methods, this work
assumes access to model parameters and does not resolve licensing, safety, or policy
compatibility issues that can arise when combining models trained under different data
and alignment constraints.

\bibliography{refs}

\appendix

\section{Additional results}

\begin{table*}[p]
  \caption{Full results across methods and merged model counts (Part 1/3).}
  \label{tab:full-results}
  \centering
  \scriptsize
  \begingroup
  \setlength{\tabcolsep}{3pt}
  \renewcommand{\arraystretch}{0.95}
  \begin{fitpage}
    \begin{sc}
     \begin{tabular}{@{}lrrrrrrr@{}}
  \toprule
  Method        & $m$ & H-Swag & BBH   & MMLU-Pro & MuSR  & GPQA-D & Avg   \\
  \midrule
  (Multi-)SLERP & 2 & 0.825 & 0.640 & 0.523 & 0.500 & 0.386 & 0.575 \\
  (Multi-)SLERP & 3 & 0.607 & 0.396 & 0.254 & 0.476 & 0.260 & 0.399 \\
  (Multi-)SLERP & 4 & 0.647 & 0.535 & 0.390 & 0.485 & 0.275 & 0.466 \\
  (Multi-)SLERP & 5 & 0.244 & 0.280 & 0.105 & 0.356 & 0.209 & 0.239 \\
  (Multi-)SLERP & 6 & 0.262 & 0.268 & 0.105 & 0.323 & 0.219 & 0.235 \\
  (Multi-)SLERP & 7 & 0.261 & 0.280 & 0.105 & 0.328 & 0.239 & 0.243 \\
  (Multi-)SLERP & 8 & 0.261 & 0.280 & 0.105 & 0.328 & 0.239 & 0.243 \\
  (Multi-)SLERP & 9 & 0.264 & 0.285 & 0.105 & 0.352 & 0.239 & 0.249 \\
  (Multi-)SLERP & 10 & 0.264 & 0.285 & 0.105 & 0.325 & 0.239 & 0.244 \\
  (Multi-)SLERP & 11 & 0.266 & 0.282 & 0.102 & 0.384 & 0.244 & 0.256 \\
  \midrule
  Arcee Fusion & 2 & 0.777 & 0.611 & 0.490 & 0.475 & 0.335 & 0.538 \\
  \midrule
  DARE-LERP & 2 & 0.770 & 0.530 & 0.418 & 0.460 & 0.325 & 0.501 \\
  DARE-LERP & 3 & 0.719 & 0.556 & 0.412 & 0.467 & 0.285 & 0.488 \\
  DARE-LERP & 4 & 0.615 & 0.505 & 0.391 & 0.489 & 0.305 & 0.461 \\
  DARE-LERP & 5 & 0.259 & 0.276 & 0.108 & 0.364 & 0.189 & 0.239 \\
  DARE-LERP & 6 & 0.258 & 0.274 & 0.106 & 0.343 & 0.209 & 0.238 \\
  DARE-LERP & 7 & 0.256 & 0.279 & 0.106 & 0.357 & 0.184 & 0.236 \\
  DARE-LERP & 8 & 0.258 & 0.293 & 0.108 & 0.333 & 0.148 & 0.228 \\
  DARE-LERP & 9 & 0.264 & 0.283 & 0.110 & 0.357 & 0.209 & 0.245 \\
  DARE-LERP & 10 & 0.259 & 0.271 & 0.107 & 0.370 & 0.219 & 0.245 \\
  DARE-LERP & 11 & 0.263 & 0.282 & 0.112 & 0.352 & 0.169 & 0.235 \\
  \midrule
  DARE-TIES & 2 & 0.767 & 0.542 & 0.434 & 0.513 & 0.330 & 0.517 \\
  DARE-TIES & 3 & 0.730 & 0.558 & 0.404 & 0.496 & 0.335 & 0.504 \\
  DARE-TIES & 4 & 0.613 & 0.467 & 0.380 & 0.440 & 0.285 & 0.437 \\
  DARE-TIES & 5 & 0.252 & 0.283 & 0.113 & 0.332 & 0.209 & 0.238 \\
  DARE-TIES & 6 & 0.256 & 0.277 & 0.108 & 0.365 & 0.219 & 0.245 \\
  DARE-TIES & 7 & 0.256 & 0.283 & 0.103 & 0.366 & 0.204 & 0.242 \\
  DARE-TIES & 8 & 0.259 & 0.284 & 0.105 & 0.340 & 0.234 & 0.244 \\
  DARE-TIES & 9 & 0.256 & 0.283 & 0.113 & 0.345 & 0.209 & 0.241 \\
  DARE-TIES & 10 & 0.259 & 0.280 & 0.107 & 0.378 & 0.229 & 0.251 \\
  DARE-TIES & 11 & 0.250 & 0.276 & 0.107 & 0.349 & 0.219 & 0.240 \\
  \bottomrule
\end{tabular}
    \end{sc}
  \end{fitpage}
  \endgroup
\end{table*}

\begin{table*}[p]\ContinuedFloat
  \caption{Full results across methods and merged model counts (Part 2/3, continued).}
  \centering
  \scriptsize
  \begingroup
  \setlength{\tabcolsep}{3pt}
  \renewcommand{\arraystretch}{0.95}
  \begin{fitpage}
    \begin{sc}
   \begin{tabular}{@{}lrrrrrrr@{}}
  \toprule
  Method     & $m$ & H-Swag & BBH   & MMLU-Pro & MuSR  & GPQA-D & Avg   \\
  \midrule
  DELLA-LERP & 2 & 0.766 & 0.547 & 0.432 & 0.492 & 0.320 & 0.511 \\
  DELLA-LERP & 3 & 0.713 & 0.554 & 0.404 & 0.525 & 0.335 & 0.506 \\
  DELLA-LERP & 4 & 0.606 & 0.498 & 0.384 & 0.475 & 0.285 & 0.450 \\
  DELLA-LERP & 5 & 0.253 & 0.281 & 0.113 & 0.345 & 0.214 & 0.241 \\
  DELLA-LERP & 6 & 0.264 & 0.284 & 0.106 & 0.361 & 0.199 & 0.243 \\
  DELLA-LERP & 7 & 0.253 & 0.283 & 0.108 & 0.361 & 0.194 & 0.240 \\
  DELLA-LERP & 8 & 0.260 & 0.274 & 0.103 & 0.361 & 0.214 & 0.242 \\
  DELLA-LERP & 9 & 0.255 & 0.287 & 0.103 & 0.373 & 0.199 & 0.243 \\
  DELLA-LERP & 10 & 0.261 & 0.280 & 0.105 & 0.341 & 0.244 & 0.246 \\
  DELLA-LERP & 11 & 0.256 & 0.291 & 0.105 & 0.345 & 0.219 & 0.243 \\
  \midrule
  DELLA-TIES & 2 & 0.765 & 0.541 & 0.426 & 0.487 & 0.325 & 0.509 \\
  DELLA-TIES & 3 & 0.783 & 0.574 & 0.445 & 0.521 & 0.355 & 0.536 \\
  DELLA-TIES & 4 & 0.779 & 0.589 & 0.475 & 0.497 & 0.381 & 0.544 \\
  DELLA-TIES & 5 & 0.266 & 0.269 & 0.111 & 0.348 & 0.244 & 0.248 \\
  DELLA-TIES & 6 & 0.262 & 0.283 & 0.105 & 0.340 & 0.214 & 0.241 \\
  DELLA-TIES & 7 & 0.253 & 0.280 & 0.114 & 0.325 & 0.199 & 0.234 \\
  DELLA-TIES & 8 & 0.264 & 0.284 & 0.107 & 0.340 & 0.239 & 0.247 \\
  DELLA-TIES & 9 & 0.255 & 0.279 & 0.109 & 0.317 & 0.244 & 0.241 \\
  DELLA-TIES & 10 & 0.259 & 0.290 & 0.108 & 0.348 & 0.229 & 0.247 \\
  DELLA-TIES & 11 & 0.260 & 0.298 & 0.108 & 0.325 & 0.229 & 0.244 \\
  \midrule
  Karcher (Ours) & 2 & 0.830 & 0.653 & 0.532 & 0.523 & 0.448 & 0.597 \\
  Karcher (Ours) & 3 & 0.833 & 0.659 & 0.538 & 0.536 & 0.443 & 0.602 \\
  Karcher (Ours) & 4 & 0.835 & 0.668 & 0.547 & 0.536 & 0.427 & 0.603 \\
  Karcher (Ours) & 5 & 0.836 & 0.680 & 0.558 & 0.532 & 0.443 & 0.610 \\
  Karcher (Ours) & 6 & 0.838 & 0.687 & 0.562 & 0.532 & 0.458 & 0.615 \\
  Karcher (Ours) & 7 & 0.833 & 0.666 & 0.518 & 0.503 & 0.448 & 0.594 \\
  Karcher (Ours) & 8 & 0.833 & 0.667 & 0.518 & 0.508 & 0.458 & 0.597 \\
  Karcher (Ours) & 9 & 0.818 & 0.666 & 0.530 & 0.519 & 0.432 & 0.593 \\
  Karcher (Ours) & 10 & 0.833 & 0.668 & 0.529 & 0.520 & 0.443 & 0.599 \\
  Karcher (Ours) & 11 & 0.835 & 0.678 & 0.538 & 0.529 & 0.443 & 0.605 \\
  \bottomrule
\end{tabular}
    \end{sc}
  \end{fitpage}
  \endgroup
\end{table*}

\begin{table*}[p]\ContinuedFloat
  \caption{Full results across methods and merged model counts (Part 3/3, continued).}
  \centering
  \scriptsize
  \begingroup
  \setlength{\tabcolsep}{3pt}
  \renewcommand{\arraystretch}{0.95}
  \begin{fitpage}
    \begin{sc}
      \begin{tabular}{@{}lrrrrrrr@{}}
        \toprule
        Method      & $m$ & H-Swag & BBH   & MMLU-Pro & MuSR  & GPQA-D & Avg   \\
        \midrule
        LERP        & 2   & 0.825  & 0.643 & 0.524    & 0.504 & 0.391  & 0.577 \\
        LERP        & 3   & 0.826  & 0.650 & 0.530    & 0.512 & 0.381  & 0.580 \\
        LERP        & 4   & 0.829  & 0.661 & 0.550    & 0.508 & 0.391  & 0.588 \\
        LERP        & 5   & 0.811  & 0.613 & 0.468    & 0.499 & 0.320  & 0.542 \\
        LERP        & 6   & 0.818  & 0.625 & 0.487    & 0.508 & 0.320  & 0.552 \\
        LERP        & 7   & 0.826  & 0.648 & 0.515    & 0.512 & 0.371  & 0.574 \\
        LERP        & 8   & 0.826  & 0.648 & 0.515    & 0.512 & 0.371  & 0.574 \\
        LERP        & 9   & 0.825  & 0.665 & 0.528    & 0.492 & 0.366  & 0.575 \\
        LERP        & 10  & 0.828  & 0.668 & 0.528    & 0.513 & 0.366  & 0.581 \\
        LERP        & 11  & 0.829  & 0.674 & 0.532    & 0.510 & 0.371  & 0.583 \\
        \midrule
        Model Stock & 3   & 0.815  & 0.610 & 0.510    & 0.459 & 0.376  & 0.554 \\
        Model Stock & 4   & 0.826  & 0.634 & 0.528    & 0.464 & 0.401  & 0.570 \\
        Model Stock & 5   & 0.823  & 0.630 & 0.508    & 0.468 & 0.355  & 0.557 \\
        Model Stock & 6   & 0.827  & 0.638 & 0.520    & 0.487 & 0.355  & 0.566 \\
        Model Stock & 7   & 0.831  & 0.650 & 0.529    & 0.493 & 0.376  & 0.576 \\
        Model Stock & 8   & 0.830  & 0.651 & 0.529    & 0.497 & 0.376  & 0.576 \\
        Model Stock & 9   & 0.827  & 0.673 & 0.538    & 0.485 & 0.396  & 0.584 \\
        Model Stock & 10  & 0.831  & 0.663 & 0.536    & 0.495 & 0.386  & 0.582 \\
        Model Stock & 11  & 0.830  & 0.670 & 0.538    & 0.495 & 0.355  & 0.578 \\
        \midrule
        SCE         & 3   & 0.804  & 0.570 & 0.471    & 0.503 & 0.355  & 0.541 \\
        SCE         & 4   & 0.801  & 0.588 & 0.479    & 0.496 & 0.366  & 0.546 \\
        SCE         & 5   & 0.735  & 0.529 & 0.342    & 0.442 & 0.260  & 0.461 \\
        SCE         & 6   & 0.715  & 0.518 & 0.306    & 0.423 & 0.255  & 0.443 \\
        SCE         & 7   & 0.715  & 0.518 & 0.305    & 0.435 & 0.260  & 0.447 \\
        SCE         & 8   & 0.714  & 0.517 & 0.305    & 0.435 & 0.260  & 0.446 \\
        SCE         & 9   & 0.628  & 0.414 & 0.266    & 0.353 & 0.265  & 0.385 \\
        SCE         & 10  & 0.716  & 0.521 & 0.310    & 0.423 & 0.260  & 0.446 \\
        SCE         & 11  & 0.716  & 0.517 & 0.311    & 0.411 & 0.270  & 0.445 \\
        \midrule
        TIES        & 2   & 0.799  & 0.590 & 0.471    & 0.505 & 0.335  & 0.540 \\
        TIES        & 3   & 0.811  & 0.595 & 0.479    & 0.505 & 0.376  & 0.553 \\
        TIES        & 4   & 0.810  & 0.615 & 0.509    & 0.505 & 0.376  & 0.563 \\
        TIES        & 5   & 0.271  & 0.288 & 0.109    & 0.357 & 0.239  & 0.253 \\
        TIES        & 6   & 0.265  & 0.287 & 0.108    & 0.360 & 0.249  & 0.254 \\
        TIES        & 7   & 0.261  & 0.279 & 0.106    & 0.346 & 0.224  & 0.243 \\
        TIES        & 8   & 0.261  & 0.279 & 0.106    & 0.346 & 0.224  & 0.243 \\
        TIES        & 9   & 0.259  & 0.292 & 0.109    & 0.336 & 0.229  & 0.245 \\
        TIES        & 10  & 0.263  & 0.265 & 0.111    & 0.354 & 0.224  & 0.243 \\
        TIES        & 11  & 0.263  & 0.275 & 0.106    & 0.331 & 0.224  & 0.240 \\
        \bottomrule
      \end{tabular}
    \end{sc}
  \end{fitpage}
  \endgroup
\end{table*}

\begin{table*}[t]
  \caption{Per-scale results grouped by \textbf{method}.}
  \label{tab:full-results-methods}
  \centering
  \scriptsize
  \begingroup
  \setlength{\tabcolsep}{3pt}
  \renewcommand{\arraystretch}{0.95}
  \begin{fitpage}
    \begin{sc}
      \begin{tabular}{llcccc}
        \toprule
        Method  & Scale & GPQA-D & H-Swag & MMLU-Pro & MuSR         \\
        \midrule
        \multicolumn{6}{l}{\textit{Methods (compared)}:}            \\
        \midrule
        Karcher & 1.7B  & 0.323  & 0.726  & 0.218    & 0.351        \\
                & 360M  & 0.268  & 0.570  & 0.120    & 0.394        \\
                & 135M  & 0.268  & 0.437  & 0.109    & 0.398        \\
        Lerp    & 1.7B  & 0.303  & 0.726  & 0.216    & 0.350        \\
                & 360M  & 0.207  & 0.568  & 0.117    & 0.395        \\
                & 135M  & 0.268  & 0.434  & 0.108    & 0.394        \\
        Slerp   & 1.7B  & 0.2929 & 0.726  & 0.218    & 0.352        \\
                & 360M  & 0.2273 & 0.569  & 0.117    & 0.398        \\
                & 135M  & 0.1919 & 0.435  & 0.109    & 0.395        \\
        Ties    & 1.7B  & 0.318  & 0.715  & 0.207    & 0.332        \\
                & 360M  & 0.258  & 0.562  & 0.114    & 0.369        \\
                & 135M  & 0.2525 & 0.428  & 0.108    & 0.390        \\
        Arcee   & 1.7B  & 0.278  & 0.719  & 0.218    & 0.343        \\
                & 360M  & 0.227  & 0.563  & 0.117    & 0.394        \\
                & 135M  & 0.207  & 0.429  & 0.113    & 0.416        \\
        TA      & 1.7B  & 0.298  & 0.718  & 0.204    & 0.341        \\
                & 360M  & 0.258  & 0.567  & 0.111    & 0.342        \\
                & 135M  & 0.273  & 0.429  & 0.109    & 0.386        \\
        \midrule
        \multicolumn{6}{l}{\textit{Reference only (not compared)}:} \\
        \midrule
        Base    & 1.7B  & 0.278  & 0.714  & 0.214    & 0.341        \\
                & 360M  & 0.247  & 0.564  & 0.116    & 0.399        \\
                & 135M  & 0.263  & 0.431  & 0.110    & 0.414        \\
        Inst    & 1.7B  & 0.298  & 0.718  & 0.204    & 0.341        \\
                & 360M  & 0.258  & 0.567  & 0.112    & 0.343        \\
                & 135M  & 0.273  & 0.429  & 0.109    & 0.386        \\
        \bottomrule
      \end{tabular}
    \end{sc}
  \end{fitpage}
  \endgroup
\end{table*}

\begin{table*}[t]
  \centering
  \small
  \begin{tabular}{l r c c c c c}
    \toprule
    Model             & Layer & Mean Variance       & EffRank        & StableRank        & PR              & NumRank        \\
    \midrule
    Qwen2.5-14B       & 0     & 0.0612 $\pm$ 0.0012 & 2284 $\pm$ 37  & 19.89 $\pm$ 0.39  & 1329 $\pm$ 16   & 4880 $\pm$ 84  \\
    Qwen2.5-14B       & 4     & 849 $\pm$ 86        & 40.4 $\pm$ 4.9 & 1.000 $\pm$ 0.000 & 2.67 $\pm$ 0.11 & 45.4 $\pm$ 4.3 \\
    Qwen2.5-14B       & 8     & 870 $\pm$ 131       & 115 $\pm$ 20   & 1.001 $\pm$ 0.000 & 4.12 $\pm$ 0.35 & 100 $\pm$ 16   \\
    Qwen2.5-14B       & 16    & 927 $\pm$ 99        & 245 $\pm$ 34   & 1.001 $\pm$ 0.000 & 6.40 $\pm$ 0.54 & 250 $\pm$ 26   \\
    Qwen2.5-14B       & 24    & 941 $\pm$ 69        & 308 $\pm$ 27   & 1.002 $\pm$ 0.000 & 7.51 $\pm$ 0.43 & 316 $\pm$ 19   \\
    Qwen2.5-14B       & 32    & 1047 $\pm$ 118      & 635 $\pm$ 67   & 1.004 $\pm$ 0.000 & 13.6 $\pm$ 1.2  & 640 $\pm$ 45   \\
    \midrule
    Karcher-14B       & 0     & 0.0766 $\pm$ 0.0018 & 2244 $\pm$ 56  & 19.45 $\pm$ 0.68  & 1299 $\pm$ 32   & 4795 $\pm$ 134 \\
    Karcher-14B       & 4     & 951 $\pm$ 53        & 42.4 $\pm$ 3.6 & 1.000 $\pm$ 0.000 & 2.70 $\pm$ 0.09 & 42.4 $\pm$ 4.0 \\
    Karcher-14B       & 8     & 1155 $\pm$ 218      & 96 $\pm$ 20    & 1.000 $\pm$ 0.000 & 3.77 $\pm$ 0.36 & 81 $\pm$ 16    \\
    Karcher-14B       & 16    & 1112 $\pm$ 188      & 211 $\pm$ 36   & 1.001 $\pm$ 0.000 & 5.80 $\pm$ 0.58 & 213 $\pm$ 29   \\
    Karcher-14B       & 24    & 1283 $\pm$ 109      & 231 $\pm$ 29   & 1.001 $\pm$ 0.000 & 6.21 $\pm$ 0.45 & 244 $\pm$ 20   \\
    Karcher-14B       & 32    & 1121 $\pm$ 200      & 633 $\pm$ 108  & 1.004 $\pm$ 0.001 & 13.5 $\pm$ 2.1  & 618 $\pm$ 94   \\
    \midrule
    Karcher-14B (reg) & 0     & 0.0765 $\pm$ 0.0013 & 2283 $\pm$ 99  & 19.91 $\pm$ 0.40  & 1322 $\pm$ 53   & 4867 $\pm$ 173 \\
    Karcher-14B (reg) & 4     & 867 $\pm$ 99        & 49.0 $\pm$ 6.4 & 1.000 $\pm$ 0.000 & 2.85 $\pm$ 0.14 & 48.8 $\pm$ 6.2 \\
    Karcher-14B (reg) & 8     & 1029 $\pm$ 152      & 110 $\pm$ 20   & 1.001 $\pm$ 0.000 & 4.02 $\pm$ 0.34 & 91 $\pm$ 13    \\
    Karcher-14B (reg) & 16    & 1214 $\pm$ 164      & 190 $\pm$ 31   & 1.001 $\pm$ 0.000 & 5.46 $\pm$ 0.48 & 195 $\pm$ 22   \\
    Karcher-14B (reg) & 24    & 1252 $\pm$ 164      & 238 $\pm$ 36   & 1.001 $\pm$ 0.000 & 6.34 $\pm$ 0.57 & 254 $\pm$ 26   \\
    Karcher-14B (reg) & 32    & 1175 $\pm$ 87       & 585 $\pm$ 41   & 1.004 $\pm$ 0.000 & 12.6 $\pm$ 0.8  & 572 $\pm$ 31   \\
    \midrule
    Ties-14B          & 0     & 0.0478 $\pm$ 0.0011 & 2090 $\pm$ 57  & 31.55 $\pm$ 0.56  & 1223 $\pm$ 35   & 4901 $\pm$ 121 \\
    Ties-14B          & 4     & 14.6 $\pm$ 1.3      & 786 $\pm$ 43   & 1.010 $\pm$ 0.001 & 22.2 $\pm$ 1.3  & 1064 $\pm$ 35  \\
    Ties-14B          & 8     & 41.2 $\pm$ 2.5      & 809 $\pm$ 43   & 1.027 $\pm$ 0.004 & 25.3 $\pm$ 1.1  & 1092 $\pm$ 35  \\
    Ties-14B          & 16    & 45.7 $\pm$ 3.1      & 887 $\pm$ 42   & 1.035 $\pm$ 0.007 & 30.1 $\pm$ 1.4  & 1209 $\pm$ 37  \\
    Ties-14B          & 24    & 49.8 $\pm$ 5.0      & 875 $\pm$ 75   & 1.044 $\pm$ 0.009 & 33.4 $\pm$ 2.7  & 1242 $\pm$ 67  \\
    Ties-14B          & 32    & 57.9 $\pm$ 4.9      & 1188 $\pm$ 69  & 1.073 $\pm$ 0.006 & 64.5 $\pm$ 5.3  & 1741 $\pm$ 83  \\
    \midrule
    Lerp-14B          & 0     & 0.0607 $\pm$ 0.0007 & 2152 $\pm$ 76  & 25.03 $\pm$ 0.24  & 1257 $\pm$ 42   & 4843 $\pm$ 148 \\
    Lerp-14B          & 4     & 104 $\pm$ 8         & 232 $\pm$ 20   & 1.002 $\pm$ 0.000 & 6.67 $\pm$ 0.30 & 317 $\pm$ 11   \\
    Lerp-14B          & 8     & 162 $\pm$ 16        & 387 $\pm$ 52   & 1.003 $\pm$ 0.000 & 9.30 $\pm$ 0.80 & 447 $\pm$ 32   \\
    Lerp-14B          & 16    & 173 $\pm$ 11        & 592 $\pm$ 48   & 1.006 $\pm$ 0.000 & 14.4 $\pm$ 1.0  & 694 $\pm$ 36   \\
    Lerp-14B          & 24    & 193 $\pm$ 24        & 717 $\pm$ 80   & 1.008 $\pm$ 0.001 & 17.9 $\pm$ 1.8  & 831 $\pm$ 58   \\
    Lerp-14B          & 32    & 186 $\pm$ 16        & 1404 $\pm$ 117 & 1.021 $\pm$ 0.002 & 46.2 $\pm$ 5.0  & 1783 $\pm$ 136 \\
    \midrule
    MS-14B            & 0     & 0.0544 $\pm$ 0.0006 & 2235 $\pm$ 31  & 22.23 $\pm$ 0.14  & 1301 $\pm$ 18   & 4891 $\pm$ 56  \\
    MS-14B            & 4     & 416 $\pm$ 46        & 68.1 $\pm$ 8.8 & 1.000 $\pm$ 0.000 & 3.32 $\pm$ 0.18 & 84.6 $\pm$ 7.5 \\
    MS-14B            & 8     & 539 $\pm$ 59        & 150 $\pm$ 19   & 1.001 $\pm$ 0.000 & 4.78 $\pm$ 0.32 & 143 $\pm$ 16   \\
    MS-14B            & 16    & 572 $\pm$ 27        & 289 $\pm$ 22   & 1.002 $\pm$ 0.000 & 7.32 $\pm$ 0.35 & 318 $\pm$ 15   \\
    MS-14B            & 24    & 581 $\pm$ 55        & 370 $\pm$ 42   & 1.003 $\pm$ 0.000 & 8.88 $\pm$ 0.72 & 406 $\pm$ 30   \\
    MS-14B            & 32    & 612 $\pm$ 45        & 825 $\pm$ 52   & 1.006 $\pm$ 0.000 & 18.6 $\pm$ 1.2  & 893 $\pm$ 42   \\
    \bottomrule
  \end{tabular}
  \caption{Layer-wise activation variance and rank diagnostics (mean $\pm$ std over bootstrap draws), where MS indicates the Model Stock method.}
  \label{tab:layerwise-rank-diagnostics}
\end{table*}

\end{document}